\pdfoutput=1

\documentclass[11pt]{article}

\usepackage[final]{acl}

\usepackage{times}
\usepackage{latexsym}

\usepackage[T1]{fontenc}

\usepackage[utf8]{inputenc}

\usepackage{microtype}

\usepackage{inconsolata}

\usepackage{graphicx}

\usepackage{color}
\usepackage[dvipsnames, svgnames]{xcolor}
\usepackage{colortbl}
\definecolor{darkblue}{rgb}{0, 0, 0.5}
\definecolor{darkgreen}{rgb}{0, 0.5, 0}

\usepackage{CJKutf8} 
\usepackage{microtype}
\usepackage{graphicx}
\usepackage{subcaption}
\usepackage{booktabs}
\usepackage{amsmath}
\usepackage{amssymb}
\usepackage{mathtools}
\usepackage{amsthm}
\usepackage{booktabs}
\usepackage{pifont}
\usepackage[capitalize,noabbrev]{cleveref}

\usepackage[utf8]{inputenc}
\usepackage[T1]{fontenc}
\usepackage{url}
\usepackage{amsfonts}
\usepackage{nicefrac}
\usepackage{latexsym}
\usepackage{array}
\usepackage{multirow}
\usepackage{alltt}
\usepackage{enumerate}
\usepackage{bbm}
\usepackage{eucal}
\usepackage{xspace}
\usepackage{stmaryrd}
\usepackage{bussproofs}
\usepackage{mathpartir}
\usepackage{tikz}
\usepackage{pgfmath}
\usepackage{xurl}
\usepackage{mdframed}
\usepackage{enumitem}
\usepackage{tabularx}
\usepackage{listings}
\usepackage{xparse}
\usepackage{xifthen}
\usepackage{multicol}
\usepackage{xfrac}
\usepackage{appendix}
\usepackage{etoolbox}
\usepackage{floatflt}
\usepackage{wrapfig}
\usepackage{tablefootnote}
\usepackage{makecell}
\usepackage{verbatim}
\usepackage{setspace}
\usepackage{floatrow}
\usepackage{adjustbox}
\usepackage{cleveref}
\usepackage{pifont}
\usepackage{longtable}
\usepackage{dashrule}
\usepackage{arydshln}
\usepackage{lipsum}
\usepackage{cleveref}
\usepackage{tcolorbox}
\usepackage{listings}
\tcbuselibrary{listingsutf8} 
\usepackage{CJKutf8}  
\usepackage{pgfplots}
\usepgfplotslibrary{groupplots}
\usepackage{caption}
\usepackage{subcaption}
\usepackage{geometry}
\usepackage{tikz}
\usetikzlibrary{patterns}
\pgfplotsset{compat=1.17}
\usetikzlibrary{patterns}
\usetikzlibrary{backgrounds}
\usepackage{amssymb}
\usepackage{times}
\usepackage{url}
\usepackage{latexsym}
\usepackage{longtable}

\usepackage{graphicx} 
\usepackage{adjustbox}
\usepackage{booktabs}
\usepackage{natbib}
\usepackage{multirow}

\usepackage{pgfplots}
\usepackage{times}
\usepackage{arydshln}

\usetikzlibrary{arrows.meta, positioning, fit, shapes.geometric, arrows}

\usepackage{ltxtable}

\usepackage{titlesec}
\titlespacing*{\paragraph}{0pt}{0.5em}{0.5em}  

\usepackage{makecell}

\usepackage{booktabs}   
\usepackage{multirow}   
\usepackage{xcolor}     
\usepackage{colortbl}   

\definecolor{tablegray}{gray}{0.9}

\usepackage{algorithm}
\usepackage[%
    noEnd=true,%
    indLines=true,%
    italicComments=false,%
    commentColor=darkgreen,%
]{algpseudocodex}

\title{RouteLMT: Learned Sample Routing for Hybrid LLM Translation Deployment}

\author{
    \textbf{Yingfeng Luo\textsuperscript{1}},
    \textbf{Hongyu Liu\textsuperscript{1}}, 
    \textbf{Dingyang Lin\textsuperscript{1}},
    \textbf{Kaiyan Chang\textsuperscript{1}},
    \textbf{Chenglong Wang\textsuperscript{1}} \\ 
    \textbf{Bei Li\textsuperscript{1}}, 
    \textbf{Quan Du\textsuperscript{2}},
    \textbf{Tong Xiao\textsuperscript{1,2}\thanks{\xspace Corresponding author.}},
    \textbf{Jingbo Zhu\textsuperscript{1,2}},
    \\
    \textsuperscript{1} School of Computer Science and Engineering, Northeastern University, Shenyang, China\\
	\textsuperscript{2} NiuTrans Research, Shenyang, China\\
    \texttt{luoyingfeng\_neu@outlook.com}\\
	\texttt{\{xiaotong,zhujingbo\}@mail.neu.edu.cn}
}

\begin{document}
\maketitle
\begin{abstract}
    Large Language Models (LLMs) have achieved remarkable performance in Machine Translation (MT), but deploying them at scale remains prohibitively expensive. 
A widely adopted remedy is the hybrid system paradigm, which balances cost and quality by serving most requests with a small model and selectively routing a fraction to a large model. 
However, existing routing strategies often rely on heuristics, external predictors, or absolute quality estimation, which fail to capture whether the large model actually provides a worthwhile improvement over the small one. 
In this paper, we formulate routing as a budget allocation problem and identify marginal gain, i.e., the large model’s improvement over the small model, as the optimal signal for budgeted decisions. 
Building on this, we propose \textbf{RouteLMT} (routing for LLM-based MT), an efficient in-model router that predicts this expected gain by probing the small translator’s prompt-token representation, without requiring external models or hypothesis decoding. 
Extensive experiments demonstrate that our RouteLMT outperforms heuristics, quality/difficulty estimation baselines, achieving a superior quality–budget Pareto frontier. 
Furthermore, we analyze regression risks and show that a simple guarded variant can mitigate severe quality losses.

\end{abstract}

\section{Introduction}
Large language models (LLMs) have recently demonstrated strong translation capability across diverse domains and language pairs \citep{DBLP:journals/corr/abs-2305-18098,DBLP:journals/corr/abs-2402-17733,DBLP:conf/iclr/Xu0SA24,DBLP:conf/naacl/CuiGLLW25,luo2025beyond}. 
However, deploying a single large model for all production requests at scale is often impractical in industrial translation systems due to strict constraints on serving cost, tail latency, and compute capacity. 
A common strategy is therefore hybrid deployment, where a smaller and cheaper model serves the majority of traffic, while a larger and more capable model is reserved for inputs where the small model is likely to translate poorly, or where the large model is expected to yield the largest quality improvements.

Such a hybrid deployment introduces a key operational question: given a fixed large model call budget, which inputs should trigger the large model?
Naive routing, such as simple heuristics (e.g., length), can waste large model capacity on easy inputs and miss high-gain cases where the large model delivers the most significant boost.
Recent work has begun to study this question (see Table~\ref{tab:related_routing}), including pre-route deciders based on source features to reduce unnecessary large model calls \citep{Wu2025CombiningTB,proietti2025estimating} and post-route, QE-based deferral strategies that first generate a small-model hypothesis translation and then use QE to defer low-quality cases to a larger model \citep{hendy2023good,farinhas2025translate}.

\newcommand{\cmark}{\ding{51}}
\newcommand{\xmark}{\ding{55}}
\begin{table}[t]
\centering
\small
\setlength{\tabcolsep}{3.5pt}
\begin{tabular}{llcc}
\toprule 
\textbf{Method} & \textbf{\makecell{Decision \\ timing}}   &  \textbf{\makecell{Needs \\hypothesis?}} & \textbf{\makecell{Extra \\ model}}  \\
\midrule
\citet{hendy2023good} & Post-route  & \cmark & \cmark  \\
\citet{Wu2025CombiningTB} & Pre-route & \xmark & \cmark   \\
\citet{farinhas2025translate} & Post-route & \cmark & \cmark   \\
\citet{proietti2025estimating}  & Pre-route & \xmark  & \cmark  \\
\textbf{RouteLMT (Ours)} & Pre-route & \xmark  & \xmark  \\
\bottomrule
\end{tabular}
\caption{Comparison of representative hybrid MT routing approaches. }
\label{tab:related_routing}
\end{table}

Despite these advances, it remains challenging to design routers that are both accurate and lightweight.
Most existing approaches either (i) rely on separate external routers (e.g., a source-feature classifier or QE model), or (ii) require generating a hypothesis from the small model before making the decision.
Both choices have clear limitations.
External routers increase operational complexity while overlooking the rich representations of the small translator, which may encode cues about input difficulty and expected translation quality.
Meanwhile, hypothesis-dependent strategies require an extra decoding and scoring step, increasing computation and latency.
Beyond overhead, many routers optimize proxies such as input difficulty or absolute quality, which are misaligned with a budgeted routing objective.
For example, hard inputs may be hard for both models, yielding zero gain, while some seemingly simple inputs might still benefit substantially from the large model.

In this work, we formulate routing as a budget allocation problem: maximizing overall translation quality under a fixed budget of large-model calls. 
This formulation motivates ranking requests by the expected \emph{marginal gain} of upgrading (the large model’s improvement over the small model) and allocating the budget on the top-ranked instances.
To predict this gain with minimal overhead, we avoid hypothesis-dependent features and instead embed the router into the small translator.
Concretely, motivated by recent findings on the informativeness of final-token representations \citep{zhu2025llm,lee2025probing,DBLP:conf/icml/DongZL0L25}, we predict gain directly from the small translator's hidden state at the last token of the translation prompt via a simple regression head. 
We jointly train this head with the small translator using parameter-efficient LoRA adaptation \citep{hu2022lora}, yielding an in-model, hypothesis-free, and translation direction-aware router without deploying a separate QE model.

To assess the efficacy of different routing policies, we systematically compare routing strategies in a budgeted hybrid LLM translation setting, spanning heuristics, learned quality/difficulty prediction, and learned gain prediction.
We evaluate routers with ranking- and allocation-oriented metrics, and summarize quality--cost trade-offs using Pareto curves that characterize translation quality as a function of the routed-to-large fraction.
Across multiple directions and budgets, we find that gain-based in-model routing delivers the best quality–cost frontier among all routers we evaluate.

In summary, our contributions are threefold:
\begin{itemize}
    \item We formulate hybrid LLM MT routing as a budget allocation problem and derive expected \emph{marginal gain} as the key signal for routing decisions.
    \item We introduce \textbf{RouteLMT}, an in-model, hypothesis-free, direction-aware router that leverages the small translator’s internal representations to predict marginal gain, without requiring external models or hypothesis decoding.
    \item We empirically validate gain-based in-model routing, showing consistent improvements over heuristics and quality/difficulty-based learned routing methods.
\end{itemize}

\section{Related Work}
\label{sec:related}

Hybrid and adaptive inference is a pragmatic way to balance quality and resource usage.
This idea has recently regained attention with LLM deployments, where the gap between model quality and serving cost/latency can be substantial.

\paragraph{Model routing and LLM cascades.}
Several works study how to route requests among multiple LLMs to improve the quality--cost frontier.
FrugalGPT \citep{DBLP:journals/tmlr/ChenZ024}, Hybrid LLM \citep{DBLP:conf/iclr/DingM0SMRLA24}, and MixLLM \citep{DBLP:conf/naacl/WangLCZCYFC25} adopt an LLM cascade perspective and study policies that selectively invoke stronger models to reduce cost while maintaining accuracy.
RouteLLM \citep{DBLP:journals/corr/abs-2406-18665} learns routers from preference data to choose between a strong and a weak LLM at inference time.
\citet{DBLP:conf/iclr/GuptaNJRMK24} systematically studies uncertainty-based deferral for generative tasks, showing that naive sequence-probability uncertainty suffers from length bias and that learned deferral rules leveraging token-level uncertainty can yield better cost--quality trade-offs.

\paragraph{Hybrid translation and quality-aware deferral.}
In machine translation, hybrid deployment has appeared in multiple forms.
Recent work combines NMT and LLM translation and learns a source-feature-based decider to reduce unnecessary LLM calls in hybrid systems \citep{Wu2025CombiningTB}.
Another prominent direction is quality-aware deferral: the system first runs a smaller translator and then uses a QE signal to decide whether to defer to a larger model.
\citet{farinhas2025translate} shows that QE-based deferral can match large-model translation quality while invoking it for only a fraction of examples.
Relatedly, \citet{hendy2023good} propose a QE-threshold approach that triggers a second (stronger) translation when the initial output is predicted to be low quality.
A complementary line of work builds predictors that estimate MT difficulty from the source sentence to identify challenging inputs \citep{DBLP:conf/emnlp/Don-YehiyaCA22,proietti2025estimating}.

Our work focuses on budgeted sample routing for hybrid LLM translation, but differs from prior MT routing in how routing signals are obtained and system design.
Rather than relying on a separately served router or post-decoding QE signals, we estimate marginal gain directly from the small translator’s internal prompt representations, without requiring a decoded hypothesis.

\section{Problem Formulation}
\label{sec:problem}
We consider sample routing for a hybrid machine translation system with a fixed budget of large-model calls.
Each request is a source sentence $x$ paired with a translation direction $d \in \mathcal{D}$ (e.g., En$\rightarrow$Zh).
The system has access to two translation models: a low-cost \emph{small} model $M_s$ and a higher-cost \emph{large} model $M_\ell$.
For a request $(x,d)$, the two models produce translations
\begin{equation}
\hat{y}_s = M_s(x; d), \qquad \hat{y}_\ell = M_\ell(x; d).
\end{equation}
A router chooses which model serves the request, resulting in the deployed output $\hat{y}$.

Let $\Phi(x,\hat{y},y^\star) \in \mathbb{R}$ denote a reference-based translation quality score, 
where $y^\star$ is the human reference translation \footnote{We use a reference-based $\Phi$ to obtain a less noisy supervision signal for isolating algorithmic gains under controlled conditions. The framework is agnostic to how $\Phi$ is obtained, and can substitute $\Phi(x,\hat{y})$ from a reference-free QE model.}.
Define the per-request quality under each model:
\begin{equation}
\begin{aligned}[t]
q_s(x; d) &= \Phi\!\left(x, \hat{y}_s, y^\star \right),\\
q_\ell(x; d) &= \Phi\!\left(x, \hat{y}_\ell, y^\star \right).
\end{aligned}
\end{equation}

We study \emph{budgeted} routing, where only a fraction $p \in (0,1]$ of requests may be served by $M_\ell$.
Let $z(x,d) \in \{0,1\}$ be the routing decision, where $z=1$ indicates routing to $M_\ell$ and $z=0$ indicates routing to $M_s$.
The budget constraint is
\begin{equation}
\label{eq:budget_constraint}
\mathbb{E}[z(x,d)] \le p,
\end{equation}
with expectation taken over the request distribution.

Under this constraint, the objective is to maximize the expected system output quality:
\begin{align}
\max_{R}\ &\mathbb{E}\Big[z(x,d) q_\ell(x; d) + (1-z(x,d)) q_s(x; d)\Big] \notag\\
\text{s.t.}\ &\mathbb{E}[z(x,d)] \le p, \label{eq:budget_objective}
\end{align}
where $R$ denotes the routing policy that maps $(x,d)$ to $z$.
It is useful to rewrite \eqref{eq:budget_objective} by defining the \emph{marginal gain} of using the large model over the small model:
\begin{equation}
\label{eq:gain}
g(x; d) = q_\ell(x; d) - q_s(x; d).
\end{equation}
Substituting \eqref{eq:gain} into \eqref{eq:budget_objective} yields
\begin{multline}
\label{eq:objective}
\mathbb{E}\Big[z\, q_\ell + (1-z)\, q_s\Big]
= \mathbb{E}\big[q_s(x;d)\big] \\
{}+ \mathbb{E}\big[z(x,d)\, g(x;d)\big].
\end{multline}

Since $\mathbb{E}[q_s(x;d)]$ is constant with respect to the router, maximizing expected system quality under a fixed call budget reduces to maximizing $\mathbb{E}[z(x,d)\, g(x;d)]$.
Therefore, the router should allocate the limited large-model budget to requests with the largest gains $g(x;d)$, i.e., prioritizing inputs where $M_\ell$ is expected to improve most over $M_s$.
This motivates routing policies that rank requests by an estimate of $g$ and allocate the budget to the top-ranked ones.
This also indicates that intuitive proxies such as difficulty or the small model’s absolute quality can be misaligned with the budgeted routing objective, as challenging inputs or low-quality small-model outputs do not necessarily imply a large improvement from the large model.

\section{Method}
\label{sec:method}

Motivated by the objective in \Cref{eq:objective}, we learn a gain predictor that estimates the marginal gain $g(x;d)=q_\ell(x;d)-q_s(x;d)$ from the input, and use it to allocate a fixed large-model budget.

\paragraph{In -model gain router.}
Let $P(x,d)$ denote the translation prompt constructed from the source sentence $x$ and direction $d$.
Given request $(x,d)$, we run a single prefill step of the small translator $M_s$ on $P(x,d)$ and extract the hidden representation of the final prompt token, denoted $h(x;d)$.
To predict the gain, we project $h(x;d)$ to a scalar score via a lightweight linear head $f_\theta$:
\begin{equation}
\hat{g}(x;d) = f_\theta\!\left(h(x;d)\right),
\end{equation}

We employ LoRA \citep{hu2022lora} to adapt the small translator $M_s$ and jointly train the regression head $f_\theta$. 
Furthermore, because the translation direction $d$ is embedded in the prompt, the representation $h(x;d)$ is inherently direction-conditioned, enabling a single router to learn direction-aware policies across multiple language pairs.

\paragraph{Supervision and training.}
For each instance $(x,y^\star,d)$, we obtain translations from $M_s$ and $M_\ell$ and compute the gain label $g(x;d)$ as the quality difference between their $\Phi$ scores.
We train the router with an MSE loss:
\begin{equation}
\mathcal{L}_{\text{gain}} = \mathbb{E}_{(x,y^\star,d)}\!\left[\left(\hat{g}(x;d) - g(x;d)\right)^2\right].
\end{equation}

\paragraph{Inference-time routing policy.}
Under a large-model budget $p$, we route requests with the highest predicted gain $\hat{g}(x;d)$.
For offline evaluation, we select the top-$p$ fraction of examples by $\hat{g}(x;d)$.
For streaming deployment, this can be implemented by applying a threshold $\tau_p$ on $\hat{g}(x;d)$, calibrated on held-out traffic, so that approximately a fraction $p$ of incoming requests satisfy $\hat{g}(x;d)\ge \tau_p$.

\begin{table*}[t]
\centering
\footnotesize 
\setlength{\tabcolsep}{3pt} 
\begin{tabular}{l c ccccc ccccc}
\toprule
\multirow{2}{*}{\textbf{Method}} & \textbf{Spearman $\uparrow$} & \multicolumn{5}{c}{\textbf{HitRate@p $\uparrow$}} & \multicolumn{5}{c}{\textbf{Mean$\Delta$@p $\uparrow$}} \\ 
\cmidrule(lr){2-2} \cmidrule(lr){3-7} \cmidrule(lr){8-12}
 & \textbf{Avg.} & En$\to$Zh & En$\to$Ru & Zh$\to$En & Ru$\to$En & \textbf{Avg.} & En$\to$Zh & En$\to$Ru & Zh$\to$En & Ru$\to$En & \textbf{Avg.} \\ \midrule
 
Gain Oracle & 1.00 & 100.00 & 100.00 & 100.00 & 100.00 & 100.00 & 17.12 & 29.36 & 10.81 & 20.64 & 19.48 \\ 
Quality Oracle  & 0.67 & 74.26 & 75.94 & 73.96 & 76.24 & 75.10 & 14.51 & 26.33 & 8.40 & 17.68 & 16.73 \\
Random & 0.00 & 30.00 & 30.00 & 30.00 & 30.00 & 30.00 & 4.90 & 10.88 & 2.25 & 5.28 & 5.83 \\
\hline 
Length & 0.24 & 47.43 & 48.32 & 47.03 & 42.77 & 46.39 & 8.50 & 17.41 & 3.76 & 7.72 & 9.35 \\
Rarity & 0.14 & 35.64 & 33.86 & 42.28 & 41.68 & 38.37 & 6.16 & 12.00 & 3.62 & 8.47 & 7.56 \\
Entropy & 0.09 & 36.93 & 38.61 & 38.61 & 34.85 & 37.25 & 6.05 & 13.29 & 3.55 & 6.89 & 7.45 \\
\hline 
sentinel-src-24 & 0.34 & 54.36 & 55.45 & \underline{54.95} & \underline{55.25} & 55.00 & \underline{9.93} & 19.49 & 4.83 & 10.83 & 11.27 \\
sentinel-src-25 & 0.31 & 50.89 & 52.57 & 51.49 & 51.49 & 51.61 & 9.36 & 18.38 & 4.57 & 10.13 & 10.61 \\
XLM-R-$\Delta$ & 0.32 & 54.06 & 55.84 & 51.68 & 52.77 & 53.59 & 9.72 & 19.49 & 4.59 & 10.29 & 11.02 \\
XLM-R-Q  & 0.28 & 50.59 & 51.78 & 50.30 & 48.22 & 50.22 & 9.18 & 18.27 & 4.40 & 9.22 & 10.27 \\
\hdashline
RouteLMT-Q & \underline{0.37} & \underline{54.95} & \underline{59.21} & \underline{54.95} & 55.05 & \underline{56.04} & 9.90 & \underline{20.47} & \underline{5.00} & \underline{11.70} & \underline{11.77} \\
\rowcolor{tablegray} \textbf{RouteLMT} & \textbf{0.40} & \textbf{56.24} & \textbf{60.50} & \textbf{55.64} & \textbf{56.93} & \textbf{57.33} & \textbf{10.24} & \textbf{21.03} & \textbf{5.12} & \textbf{12.13} & \textbf{12.13} \\

\bottomrule
\end{tabular}
\caption{Router performance across four translation directions. Spearman is computed over the full set to measure global ranking quality, while HitRate@p and Mean$\Delta$@p are evaluated under a fixed budget $p{=}0.3$ (30\% routed to the large model). \textbf{Bold} numbers indicate the best score, and \underline{underlined} numbers the second best.}
\label{tab:router_perf}
\end{table*}

\section{Experiments}
\label{sec:exp}

\subsection{Data and Models}
We evaluate four translation directions: En$\leftrightarrow$Zh and En$\leftrightarrow$Ru.
For the training, we use the \textit{General Translation} split of ComMT \citep{DBLP:conf/acl/LuoZMLZGXFLXZ25}, a compiled dataset aggregating high-quality parallel corpora from multiple sources.
For evaluation, we combine FLORES-200 devtest \citep{DBLP:journals/corr/abs-2207-04672}, WMT24++ \citep{deutsch2025wmt24expandinglanguagecoverage}, and BOUQuET \citep{andrews2025bouquet}.
Detailed statistics are provided in Table~\ref{tab:data}.

We employ the recently released open LMT-60 multilingual MT model family \citep{luo2025beyond}, which provides models across a range of sizes. 
We choose LMT-60-0.6B as $M_s$ and LMT-60-8B as $M_\ell$ as a representative small/large configuration for hybrid routing.
We apply LoRA to all linear layers of $M_s$ with rank $r{=}8$ and $\alpha{=}32$; full hyperparameter settings are in Appendix Table~\ref{tab:hyperparameters}.

\subsection{Baselines}
\label{sec:baselines}
We compare RouteLMT against random routing, heuristic routers, and learned routers.
Given a large-model budget $p$, each method assigns a routing score to each request and routes approximately the top-$p$ fraction to $M_\ell$ (or bottom-$p$ when lower scores indicate higher priority).

Heuristic routers include \textbf{Length}, \textbf{Rarity}, and \textbf{Entropy}.
Learned routers are grouped by router type and learning target.
In terms of router type, we distinguish between (i) in-model routers that leverage the small translator’s internal representations and (ii) external routers that rely on a separate model operating on the source sentence.
In terms of learning target, we consider predicting marginal gain ($\Delta$) versus small-model quality ($Q$).
We denote our proposed in-model gain router as \textbf{RouteLMT} (i.e., RouteLMT-$\Delta$), and its quality variant as \textbf{RouteLMT-Q}.
As external counterparts, we train \textbf{XLM-R-$\Delta$} and \textbf{XLM-R-Q} using XLM-RoBERTa-L \citep{DBLP:conf/acl/ConneauKGCWGGOZ20} on the same data as RouteLMT to predict $\Delta$ and $Q$ from the source sentence.
Additionally, we include \textbf{sentinel-src-24/25} \citep{proietti2025estimating}, a strong baseline that models translation difficulty via quality estimation.
A summary is provided in Table~\ref{tab:baseline_details}, with full details in Appendix~\ref{app:baselines}.

\subsection{Metrics}
We adopt XCOMET-XXL \citep{guerreiro2024xcomet} as the reference-based translation quality evaluator $\Phi$, which has been shown to achieve high agreement with human judgments.
We evaluate routing performance using three complementary metrics:

\begin{itemize}[noitemsep]
    \item \textbf{Spearman}: The rank correlation between predicted routing scores and the oracle marginal gain. This assesses the router's ability to induce a globally consistent ranking of improvement potential.
    \item \textbf{HitRate@$p$}: The overlap fraction between the router-selected top-$p$ set and the oracle top-$p$ set. This evaluates selection accuracy (or decision accuracy) under a budget $p$.
    \item \textbf{Mean$\Delta$@$p$}: The average marginal gain of the router-selected top-$p$ examples. This measures the actual benefit of the routed subset.
\end{itemize}

\section{Results and Analyses}

\begin{figure*}[ht]
\centering
\includegraphics[width=\linewidth]{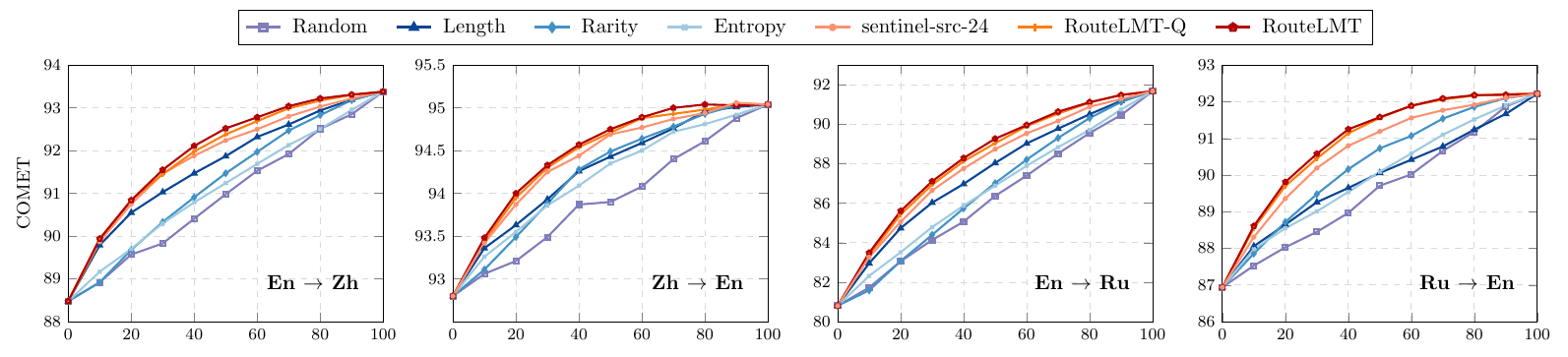}
\caption{Quality--budget trade-offs of hybrid translation routing. We sweep the large-model budget $p$ (route-to-large rate) and report the quality of the resulting hybrid system. Higher curves indicate a better Pareto frontier.}
\label{fig:pareto}
\end{figure*}

\subsection{Router Performance}
\label{sec:router_perf}

Table~\ref{tab:router_perf} reports router performance under a fixed large-model budget $p{=}0.3$, which we use as a representative operating point for practical quality--cost trade-off.
We include two oracle baselines to contextualize the upper bounds.
\textbf{Gain Oracle} routes by the true marginal gain $g(x;d)$ and is therefore optimal under the budget constraint, whereas \textbf{Quality Oracle} routes by the true small-model quality, which requires first decoding the small-model hypothesis and then scoring it, and thus serves as an upper bound for quality-based strategy.
A key takeaway from the oracles is that maximizing absolute quality is not equivalent to maximizing marginal gain: even with oracle access, quality-based routing is only moderately correlated with true gain (Spearman: 0.67) and consequently underperforms the Gain Oracle in both HitRate and Mean$\Delta$.

Among practical routers, RouteLMT (in-model gain prediction) performs best overall across all three metrics.
It achieves the highest Spearman correlation and HitRate@$p$, indicating a consistently better global ranking and top-$p$ selection capabilities than compared methods.
More importantly, it yields the largest Mean$\Delta$@$p$ (12.13), improving over the strongest heuristic Length baseline (9.35) by +2.78 and more than doubling the Random baseline (5.83).
When comparing learning objectives, RouteLMT consistently surpasses its quality-predicting counterpart (RouteLMT-Q), confirming that directly modeling marginal gain aligns better with the budgeted allocation goal.
Furthermore, external routers (XLM-R-$\Delta$/Q and sentinel-src-24/25) consistently trail the in-model predictors (RouteLMT and RouteLMT-Q), highlighting the value of leveraging the small translator’s internal representations for learning routing policies.
We also observe these trends generalize to out-of-domain settings (medical and colloquial); detailed results are reported in Appendix~\Cref{sec:domain}.

Overall, these results show that effective routing benefits from signals encoded in the small translator’s prompt representations, and that predicting marginal gain is more effective than relying on quality/difficulty proxies. 
Despite these improvements, the remaining gap to the \textbf{Gain Oracle} highlights substantial headroom for future improvements in learned routing.

\subsection{Quality--Budget Pareto Frontier}
\label{sec:pareto}

To characterize router behavior beyond a single operating point, we sweep the large-model budget $p$ and visualize the resulting quality--budget curves in Figure~\ref{fig:pareto}.
For clarity, we plot representative policies from each family (random, heuristics, and learned routers), and include sentinel-src-24 as the strongest external predictor.

Figure~\ref{fig:pareto} shows that learned routers (red family) consistently dominate heuristic policies (blue family) across directions.
This indicates that routing benefits from learned decision signals beyond surface heuristics such as length- or rarity-based proxies.
A secondary pattern is diminishing returns as $p$ increases: once the highest-return requests have been routed, additional large-model calls yield smaller quality gains, so the curves gradually converge near full routing.

Among learned routers, RouteLMT achieves the best quality--budget Pareto frontier across directions over most budgets. 
In addition, RouteLMT-Q performs better than the external predictor sentinel-src-24, indicating that leveraging the internal representations of the translation backbone provides a more informative routing signal than a stand-alone model operating solely on source text.

\subsection{Risk Analysis}
\label{sec:risk}

In deployment, improving the average is not enough; practitioners also care about regression risk and their operational impact.
We therefore analyze the gain distribution of routed-to-large requests under the same budget ($p{=}0.3$).
For each routed example, we compute $g=\Phi(\hat{y}_\ell)-\Phi(\hat{y}_s)$ and bucket it into Severe loss ($g\le -5$), Minor loss ($-5<g<-0.5$), Tie ($|g|\le 0.5$), and Substantial gain ($g>0.5$).
Figure~\ref{fig:risk} reports bucket proportions averaged across four directions (per-direction results in Appendix Figure~\ref{fig:risk_all}).

Figure~\ref{fig:risk} shows that learned routing reallocates the large-model budget toward clearly beneficial upgrades.
Compared to random and simple heuristics, learned routers markedly reduce \emph{Tie} cases, where large-model calls yield little benefit, and correspondingly increase the share of \emph{Substantial gain} invocations.
Among practical methods, RouteLMT achieves the highest substantial-gain proportion while keeping minor-loss rates low, indicating a more efficient use of the fixed budget.
Notably, severe-loss rates remain non-trivial and are similar across most non-random policies (roughly 8--9\%), suggesting that the main gains of learned routing come from prioritizing high-return requests rather than eliminating extreme negative-gain cases.

\subsection{Guarded Routing for Risk Control}
\label{sec:guarded}

To further mitigate severe regressions, we explore a simple guarded variant that augments gain-based routing with an additional quality filter.
Concretely, we first rank requests by predicted gain (RouteLMT) and then apply a quality guard, while keeping the overall route-to-large rate fixed at $p{=}0.3$ to ensure a fair comparison.
We set the guard threshold using a fixed quantile (30\% in our experiments) as a simple calibration choice.
We consider two guards: Quality (predict) uses an in-model quality predictor (RouteLMT-Q), while Quality (hypo) first decodes the $M_s$ hypothesis and then applies a quality scorer.

Table~\ref{tab:guard} shows that Quality (predict) has little effect in this setting, yielding nearly identical severe-loss and Mean$\Delta$@$p$ to gain-only routing.
In contrast, Quality (hypo) substantially reduces severe loss (from 8.19\% to 5.69\%) and improves Mean$\Delta$@$p$ (from 12.13 to 16.73), suggesting that a portion of catastrophic cases can be detected given a post-route verifier.

Overall, these results suggest a coarse-to-fine scheme: use gain-based pre-routing to handle most requests with low overhead, and optionally apply a post-route check on the remaining candidates when stricter risk control is needed, at the cost of additional decoding and scoring.

\begin{figure}[t]
\centering
\includegraphics[width=\linewidth]{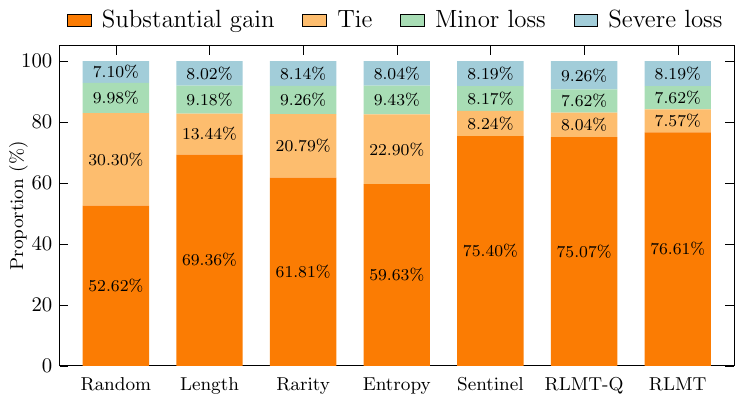}
\caption{Gain-bucket distribution among routed-to-large model requests under budget $p{=}0.3$, averaged across four directions.}
\label{fig:risk}
\end{figure}

\begin{table}[t]
\centering
\small
\setlength{\tabcolsep}{6pt}
\begin{tabular}{lcc}
\toprule
\textbf{Method} & \textbf{Severe loss} $\downarrow$ & \textbf{Mean$\Delta@$p} $\uparrow$ \\
\midrule
Random & 7.10\% & 5.83 \\
Gain & 8.19\% & 12.13 \\
Gain + Quality (predict) & 8.19\% & 12.24 \\
Gain + Quality (hypo) & 5.69\% & 16.73 \\
\bottomrule
\end{tabular}
\caption{Guarded routing ($p{=}0.3$) with gain ranking and quality guards.}
\label{tab:guard}
\end{table}

\newcommand{\hl}[1]{\colorbox{gray!15}{#1}} 
\newcolumntype{Y}{>{\raggedright\arraybackslash}X}
\begin{table*}[!t]
\begin{CJK*}{UTF8}{gbsn}
\centering
\footnotesize
\setlength{\tabcolsep}{3pt}
\renewcommand{\arraystretch}{1.15}
\begin{tabularx}{\columnwidth}{l l l Y r}
\toprule
Case & Type & Item & Text & Score \\
\midrule

\multirow{4}{*}{\makecell[l]{En$\rightarrow$Zh\\$\Delta=+25.36$}}
& \multirow{4}{*}{\makecell[l]{Code\\-switch}}
& Source   & I'm about halfway thru the first study unit & -- \\
&  & Reference   & 第一个学习单元我已经学完一半了 & -- \\
&  & Small & 我大约 \hl{halfway} 到了第一个学习单元 & 74.59 \\
&  & Large & 我已经完成了第一个学习单元的一半内容 & 99.95 \\
\midrule

\multirow{4}{*}{\makecell[l]{En$\rightarrow$Zh\\$\Delta=+25.42$}}
& \multirow{4}{*}{\makecell[l]{Literal}}
& Source   & we understand that it is hard to get to these positions of wealth and power without stepping on others along the way. & -- \\
&  & Reference   & 我们逐渐意识到，一个人如果想要获得财富和权力，就很难保证自己不会去伤害别人、利用别人。 & -- \\
&  & Small & 我们明白，要想获得财富和权力的这些地位，就必须在途中\hl{踩到别人的脚}。 & 72.86 \\
&  & Large & 我们知道，如果不踩着别人往上爬，就很难获得财富和权力。 & 98.28 \\
\midrule

\multirow{4}{*}{\makecell[l]{En$\rightarrow$Zh\\$\Delta=-10.10$}}
& \multirow{4}{*}{\makecell[l]{Entity}}
& Source   & Bring back oversight for WA's jails. Lives depend on it & -- \\
&  & Reference   & 恢复对华盛顿州看守所的监管刻不容缓，人命关天！ & -- \\
&  & Small & 恢复 \hl{WA} 监狱的监管。生命取决于此 & 83.60 \\
&  & Large & 恢复对\hl{西澳大利亚州}监狱的监管。生命攸关 & 73.50 \\
\midrule

\multirow{4}{*}{\makecell[l]{Zh$\rightarrow$En\\$\Delta=-36.33$}}
& \multirow{4}{*}{\makecell[l]{Over-\\paraphrase}}
& Source   & 我喜欢说什么就说什么。 & -- \\
&  & Reference   & I can say whatever I want. & -- \\
&  & Small & I like to say whatever I want to say. & 93.52 \\
&  & Large & I like to \hl{speak my mind}. & 57.19 \\
\bottomrule
\end{tabularx}

\caption{Case study examples with highlighted error spans. $\Delta=q_L-q_S$ is computed from XCOMET-XXL scores (scaled to 0--100). Type summarizes the primary error type.}
\label{tab:case_study}
\end{CJK*}
\end{table*}

\subsection{Case Study}
\label{sec:case_study}

Table~\ref{tab:case_study} provides illustrative examples showing both large positive gains and negative-gain regressions when upgrading to $M_\ell$.
These cases offer qualitative context for the regression patterns discussed in \S\ref{sec:risk}.

In the positive-gain examples, the large model mainly corrects systematic weaknesses of the small translator.
For En$\rightarrow$Zh, $M_s$ occasionally fails on code-switching and leaves English fragments untranslated (e.g., “halfway”), whereas $M_\ell$ produces a complete and fluent Chinese rendering ($\Delta{=}+25.36$).
We also observe that $M_\ell$ better handles figurative or non-literal expressions: when $M_s$ translates a metaphor too literally, $M_\ell$ produces a more idiomatic formulation with improved adequacy and fluency ($\Delta{=}+25.42$).

The negative-gain examples highlight plausible regression modes.
One involves abbreviation handling. 
In the headline example, $M_\ell$ attempts to interpret “WA” and expands it to a specific entity, but chooses the wrong one (“Western Australia”), leading to a clear adequacy error. 
By contrast, $M_s$ largely preserves “WA” as-is, which is less informative but avoids committing to an incorrect expansion; under reference-based scoring, the large model is penalized more heavily for the wrong disambiguation ($\Delta{=}-10.10$).
A second class involves semantic drift under paraphrasing.
For Zh$\rightarrow$En, $M_\ell$ generates a more idiomatic paraphrase (“speak my mind”) that shifts meaning relative to the intended sense (“say whatever I want”), resulting in a large negative gain ($\Delta{=}-36.33$).
Although reference-based metrics can sometimes penalize benign paraphrases, here the large-model output also shifts the intended meaning, so the regression is not merely stylistic.

Taken together, these cases suggest that many severe regressions arise from rare but high-impact generation errors (e.g., wrong entity resolution or meaning drift) rather than from the same “difficulty” factors that make $M_s$ weak.
This helps explain why severe-loss rates can remain similar across many routing policies.

\section{Conclusion}
In this paper, we studied learned sample routing for hybrid LLM translation deployment. 
By formulating routing as a budget allocation problem, we identify marginal gain as the key decision signal and propose RouteLMT, an in-model router built on the small translator that predicts gain from its translation-prompt representations using lightweight LoRA adaptation.
Across four translation directions, gain-based in-model routing achieves the strongest quality--budget trade-off among evaluated baselines, outperforming heuristic policies and separately trained routers that operate solely on source text.
We further analyzed gain-bucket distributions to characterize regression risk, and showed that a simple guarded variant can reduce severe-loss cases. 
Overall, our results suggest that probing small-model representations offers an effective approach to budgeted hybrid translation routing.

\section*{Limitations}
Our study provides an initial validation of the approach, and we note several limitations.
First, the gain supervision is derived from an automatic reference-based quality metric, which may not fully reflect human judgments or application-specific utility, and the learned router can inherit its biases. 
Second, we focus on a two-model hybrid setting with a fixed route-to-large budget; more complex cascades, multi-tier budgets, and latency-aware objectives remain to be explored. 
In addition, our experiments use a specific small/large pair (0.6B vs.\ 8B), a representative small/large configuration; how the same routing design behaves with other model families, larger scales, or different capability gaps remains to be validated.
Finally, our experiments cover four directions (En$\leftrightarrow$Zh and En$\leftrightarrow$Ru); broader language coverage and large-scale multilingual settings, where routing behavior may vary across scripts and resource levels, are important directions for future work.

\section*{Acknowledgments}
This work was supported in part by the National Natural Science Foundation of China (Nos. U24A20334 and 62276056), the Yunnan Fundamental Research Projects (No.202401BC070021), the Yunnan Science and Technology Major Project (No. 202502AD080014), the Fundamental Research Funds for the Central Universities (Nos. N25BSS054 and N25BSS094), and the Program of Introducing Talents of Discipline to Universities, Plan 111 (No.B16009).


\bibliography{references}


{
\appendices
\clearpage

\section{Baseline Details}
\label{app:baselines}

We compare against a diverse set of routing baselines, including random routing, heuristics, and learned routers.
All routing methods output a scalar routing score $s$ from the source sentence $x$ (and direction $d$ when applicable).
Given a large-model budget $p$, we route the top-$p$ (or bottom-$p$) fraction of requests by $s$ to $M_\ell$ and the remainder to $M_s$.

\begin{itemize}[leftmargin=*, itemsep=2pt, topsep=2pt, parsep=0pt, partopsep=0pt]
  \item \textbf{Random.} Routes a uniformly random $p$ fraction of requests to $M_\ell$.
  \item \textbf{Heuristic routers.}
  \begin{itemize}[leftmargin=*, itemsep=1pt, topsep=1pt, parsep=0pt, partopsep=0pt]
    \item \textbf{Length:} Sentence length under tokenization (\texttt{spaCy} for En/Ru; \texttt{jieba} for Zh), with longer inputs prioritized.
    \item \textbf{Rarity:} We compute unigram frequencies $f(w)$ from \texttt{wordfreq} and score each sentence by the average surprisal $-\log f(w)$ of its bottom 30\% least-frequent tokens (higher $\Rightarrow$ higher routing priority).
    \item \textbf{Entropy:} Entropy of $M_s$'s next-token distribution at the first decoding step, used as a lightweight uncertainty proxy (higher $\Rightarrow$ higher routing priority).
  \end{itemize}

 \item \textbf{Learned routers.} We consider learned routing with two system forms: \emph{in-model} routers that leverage the small translator's representations, and \emph{external} routers based on separately served encoder-only models. Each form is studied under two learning targets: predicting marginal gain ($\Delta$) or predicting small-model quality ($Q$).
  \begin{itemize}[leftmargin=*, itemsep=1pt, topsep=1pt, parsep=0pt, partopsep=0pt]
    \item \textbf{In-model routers.}
    \begin{itemize}[leftmargin=*, itemsep=1pt, topsep=1pt, parsep=0pt, partopsep=0pt]
      \item \textbf{RouteLMT ($\Delta$; ours).} The in-model gain router in \S\ref{sec:method} that predicts marginal gain.
      \item \textbf{RouteLMT-Q.} Same architecture as RouteLMT, trained to predict the small model's quality.
    \end{itemize}

    \item \textbf{External encoder-only routers.}
    \begin{itemize}[leftmargin=*, itemsep=1pt, topsep=1pt, parsep=0pt, partopsep=0pt]
      \item \textbf{XLM-R-$\Delta$.} An XLM-RoBERTa-Large model trained on the same supervision as RouteLMT to predict marginal gain from the source.
      \item \textbf{XLM-R-Q.} An XLM-RoBERTa-Large model trained on the same supervision to predict the small model's quality from the source.
      \item \textbf{sentinel-src-24/25} \citep{proietti2025estimating}. Publicly released XLM-RoBERTa-based source-only model trained for translation quality prediction.
    \end{itemize}
  \end{itemize}
\end{itemize}

\begin{table}[t]
\centering
\begin{tabularx}{\linewidth}{>{\raggedright\arraybackslash}X r}
\hline
\textbf{Hyperparameter} & \textbf{Value} \\
\hline
Train Type & LoRA \\
Learning Rate & 1e-4 \\
Number of Epochs & 1 \\
Global Batch Size & 64 \\
Max Length & 1024 \\ 
Warmup Ratio & 0.05 \\
LoRA Rank & 8 \\
LoRA Alpha & 32 \\
Target Modules & all-linear \\
Precision & bfloat16 \\
\hline
\end{tabularx}
\caption{Hyperparameter configuration for the regression task.}
\label{tab:hyperparameters}
\end{table}

\begin{table}[t]
\centering
\small
\begin{tabularx}{\columnwidth}{l X c r}
\toprule
Split & Source & Direction & \# Size \\
\midrule
\multirow{2}{*}{Train} &
\multirow{2}{=}{ComMT (General Translation)} &
En$\leftrightarrow$Zh & 56{,}918 \\
& & En$\leftrightarrow$Ru & 40{,}862 \\
\midrule
\multirow{2}{*}{Eval} &
\multirow{2}{=}{FLORES + BOUQuET + WMT24++} &
En$\leftrightarrow$Zh & 3{,}368 \\
& & En$\leftrightarrow$Ru & 3{,}368 \\
\bottomrule
\end{tabularx}
\caption{Training and evaluation datasets used in our experiments.}
\label{tab:data}
\end{table}

\begin{table*}[t]
\centering
\small
\setlength{\tabcolsep}{4pt}
\begin{tabular}{lcccccc}
\toprule
\textbf{Method} & \textbf{Type} & \textbf{Form} & \textbf{Target} & \textbf{Score} & \textbf{Route} & \textbf{Extra model} \\
\midrule
Random & -- & -- & -- & uniform & top-$p$ & \xmark \\
\midrule
Length  & Heuristic & -- & -- & length & top-$p$ & \xmark \\
Rarity  & Heuristic & -- & -- & surprisal & top-$p$ & \xmark \\
Entropy & Heuristic & -- & -- & entropy & top-$p$ & \xmark \\
\midrule
XLM-R-$\Delta$ & Learned & External & $\Delta$ & $\hat{g}(x;d)$ & top-$p$ & \cmark \\
XLM-R-Q & Learned & External & $Q$ & $\hat{q}_s(x;d)$ & bottom-$p$ & \cmark \\
sentinel-src-24/25 \citep{proietti2025estimating} & Learned & External & $Q$ & $\hat{q}_s(x;d)$ & bottom-$p$ & \cmark \\
\midrule
RouteLMT ($\Delta$; ours) & Learned & In-model & $\Delta$ & $\hat{g}(x;d)$ & top-$p$ & \xmark \\
RouteLMT-Q & Learned & In-model & $Q$ & $\hat{q}_s(x;d)$ & bottom-$p$ & \xmark \\
\bottomrule
\end{tabular}
\caption{Baselines for budgeted hybrid routing. Route indicates whether we select the top-$p$ or bottom-$p$ fraction under budget $p$, depending on the score semantics. $\Delta$ denotes marginal-gain prediction and $Q$ denotes small-model quality/difficulty prediction.}
\label{tab:baseline_details}
\end{table*}

\section{Domain Generalization}
\label{sec:domain}

In practice, a routing policy is exposed to a wide mix of user inputs, and the input distribution can across domains, so robustness to domain shift is important.
To test robustness, we reuse the same routers from Table~\ref{tab:router_perf}, trained on the ComMT General Translation split.
We then evaluate them on medical (Table~\ref{tab:router_perf_medical}) and colloquial (Table~\ref{tab:router_perf_colloquial}) domain from the ComMT \citep{DBLP:conf/acl/LuoZMLZGXFLXZ25} domain test sets without any additional adaptation.
We evaluate under the same budgeted setting with $p{=}0.3$.

The overall ordering of methods is consistent with the main results in Table~\ref{tab:router_perf}. 
On both domains, gain-based in-model routing remains the strongest practical approach. 
In particular, RouteLMT yields higher allocation benefit than surface heuristics and is competitive with or better than external source-only predictors. 
Ranking and selection metrics show the same pattern, indicating that the learned routing signal transfers under domain shift.
These results indicate that routing signals derived from the small translator’s prompt representations transfer well across domains.

\begin{table*}[t]
\centering
\footnotesize
\setlength{\tabcolsep}{3pt}
\begin{tabular}{l c ccccc ccccc}
\toprule
\multirow{2}{*}{\textbf{Method}} & \textbf{Spearman $\uparrow$} & \multicolumn{5}{c}{\textbf{HitRate@p $\uparrow$}} & \multicolumn{5}{c}{\textbf{Mean$\Delta$@p $\uparrow$}} \\ 
\cmidrule(lr){2-2} \cmidrule(lr){3-7} \cmidrule(lr){8-12}
 & \textbf{Avg.} & En$\to$Zh & En$\to$Ru & Zh$\to$En & Ru$\to$En & \textbf{Avg.} & En$\to$Zh & En$\to$Ru & Zh$\to$En & Ru$\to$En & \textbf{Avg.} \\ \midrule
 
Gain Oracle & 1.00 & 100.00 & 100.00 & 100.00 & 100.00 & 100.00 & 12.55 & 25.36 & 8.30 & 13.53 & 14.94 \\ 
Quality Oracle & 0.64 & 69.36 & 77.24 & 74.89 & 74.58 & 74.02 & 9.88 & 22.92 & 7.06 & 11.13 & 12.75 \\
Random & 0.00 & 30.00 & 30.00 & 30.00 & 30.00 & 30.00 & 3.00 & 9.27 & 1.95 & 3.04 & 4.32 \\
\hline
Length & 0.18 & 39.29 & 42.86 & 44.40 & 39.71 & 41.56 & 4.36 & 13.63 & 3.22 & 4.64 & 6.46 \\
Rarity & 0.14 & 32.20 & 42.86 & 38.58 & 40.92 & 38.64 & 3.30 & 12.99 & 2.68 & 5.07 & 6.01 \\
Entropy & 0.05 & 33.62 & 32.93 & 34.04 & 32.69 & 33.32 & 3.69 & 10.66 & 2.31 & 4.17 & 5.21 \\
\hline
sentinel-src-24 & 0.26 & 42.55 & 49.64 & 47.38 & 49.88 & 47.36 & 4.99 & 15.21 & 3.78 & 6.43 & 7.60 \\
sentinel-src-25 & 0.24 & 43.40 & 48.91 & 46.95 & 48.43 & 46.92 & 5.07 & 15.17 & 3.71 & 5.90 & 7.46 \\
XLM-R-$\Delta$ & 0.23 & 40.28 & 45.76 & 46.81 & 47.22 & 45.02 & 4.79 & 14.44 & 3.53 & 5.82 & 7.15 \\
XLM-R-Q & 0.18 & 39.72 & 43.34 & 45.67 & 41.89 & 42.66 & 4.81 & 13.86 & 3.41 & 5.09 & 6.79 \\
\hdashline
RouteLMT-Q & \underline{0.30} & \underline{46.52} & \underline{53.03} & \underline{48.94 } & \textbf{51.57} & \underline{51.04} & \textbf{5.84} & \underline{16.43} & \underline{3.99} & \textbf{6.69} & \underline{8.24} \\
\rowcolor{tablegray}
\textbf{RouteLMT} & \textbf{0.30} & \textbf{47.66} & \textbf{54.72} & \textbf{53.03} & \underline{50.12} & \textbf{51.39} & \underline{5.81} & \textbf{16.78} & \textbf{4.39} & \underline{6.64} & \textbf{8.41} \\
\bottomrule
\end{tabular}
\caption{Router performance on the \textit{medical} domain across four translation directions. Spearman measures global ranking quality, while HitRate@p and Mean$\Delta$@p are evaluated with a fixed budget $p{=}0.3$. \textbf{Bold} indicates the best result and \underline{underline} the second best.}
\label{tab:router_perf_medical}
\end{table*}

\begin{table*}[t]
\centering
\footnotesize
\setlength{\tabcolsep}{3pt}
\begin{tabular}{l c ccccc ccccc}
\toprule
\multirow{2}{*}{\textbf{Method}} & \textbf{Spearman $\uparrow$} & \multicolumn{5}{c}{\textbf{HitRate@p $\uparrow$}} & \multicolumn{5}{c}{\textbf{Mean$\Delta$@p $\uparrow$}} \\ 
\cmidrule(lr){2-2} \cmidrule(lr){3-7} \cmidrule(lr){8-12}
 & \textbf{Avg.} & En$\to$Zh & En$\to$Ru & Zh$\to$En & Ru$\to$En & \textbf{Avg.} & En$\to$Zh & En$\to$Ru & Zh$\to$En & Ru$\to$En & \textbf{Avg.} \\ \midrule
 
Gain Oracle & 1.00 & 100.00 & 100.00 & 100.00 & 100.00 & 100.00 & 7.77 & 28.51 & 7.30 & 17.22 & 15.20 \\ 
Quality Oracle & 0.61 & 69.80 & 66.98 & 74.07 & 80.06 & 72.73 & 6.11 & 24.36 & 6.35 & 14.96 & 12.94 \\
Random & 0.00 & 30.00 & 30.00 & 30.00 & 30.00 & 30.00 & 1.49 & 11.84 & 1.38 & 4.74 & 4.86 \\
\hline
Length & 0.21 & 45.30 & 45.79 & 45.87 & 38.01 & 43.74 & 3.30 & 17.17 & 2.16 & 6.13 & 7.19 \\
Rarity & 0.14 & 33.90 & 36.45 & 42.17 & 41.12 & 38.41 & 1.99 & 14.15 & 2.85 & 6.88 & 6.47 \\
Entropy & 0.03 & 32.76 & 31.46 & 36.75 & 35.83 & 34.20 & 1.54 & 13.71 & 2.28 & 5.64 & 5.79 \\
\hline
sentinel-src-24 & 0.26 & 44.73 & 48.60 & 52.71 & 51.09 & 49.28 & 3.12 & 17.72 & 2.76 & 8.72 & 8.08 \\
sentinel-src-25 & 0.21 & 43.59 & 43.93 & 48.43 & 45.48 & 45.36 & 2.75 & 16.36 & 3.19 & 7.32 & 7.40 \\
XLM-R-$\Delta$ & 0.24 & 49.00 & 47.66 & 45.01 & 49.84 & 47.88 & 3.52 & 17.47 & 2.21 & 9.15 & 8.09 \\
XLM-R-Q & 0.19 & 47.86 & 46.11 & 42.45 & 44.86 & 45.32 & 3.44 & 16.66 & 1.84 & 7.62 & 7.39 \\
\hdashline
RouteLMT-Q & \underline{0.29} & \textbf{49.00} & \underline{48.91} & \underline{51.85} & \textbf{53.89} & \underline{50.91} & \underline{3.48} & \underline{18.17} & \underline{3.86} & \textbf{10.19} & \textbf{8.92} \\
\rowcolor{tablegray}
\textbf{RouteLMT} & \textbf{0.30} & \underline{48.72} & \textbf{50.47} & \textbf{52.99} & \underline{52.96} & \textbf{51.28} & \textbf{3.68} & \textbf{18.50} & \textbf{3.90} & \underline{9.41} & \underline{8.87} \\
\bottomrule
\end{tabular}
\caption{Router performance on the \textit{colloquial} domain across four translation directions. Spearman measures global ranking quality, while HitRate@p and Mean$\Delta$@p are evaluated with a fixed budget $p{=}0.3$. \textbf{Bold} indicates the best result and \underline{underline} the second best.}
\label{tab:router_perf_colloquial}
\end{table*}

\begin{figure*}[ht]
\centering
\includegraphics[width=\linewidth]{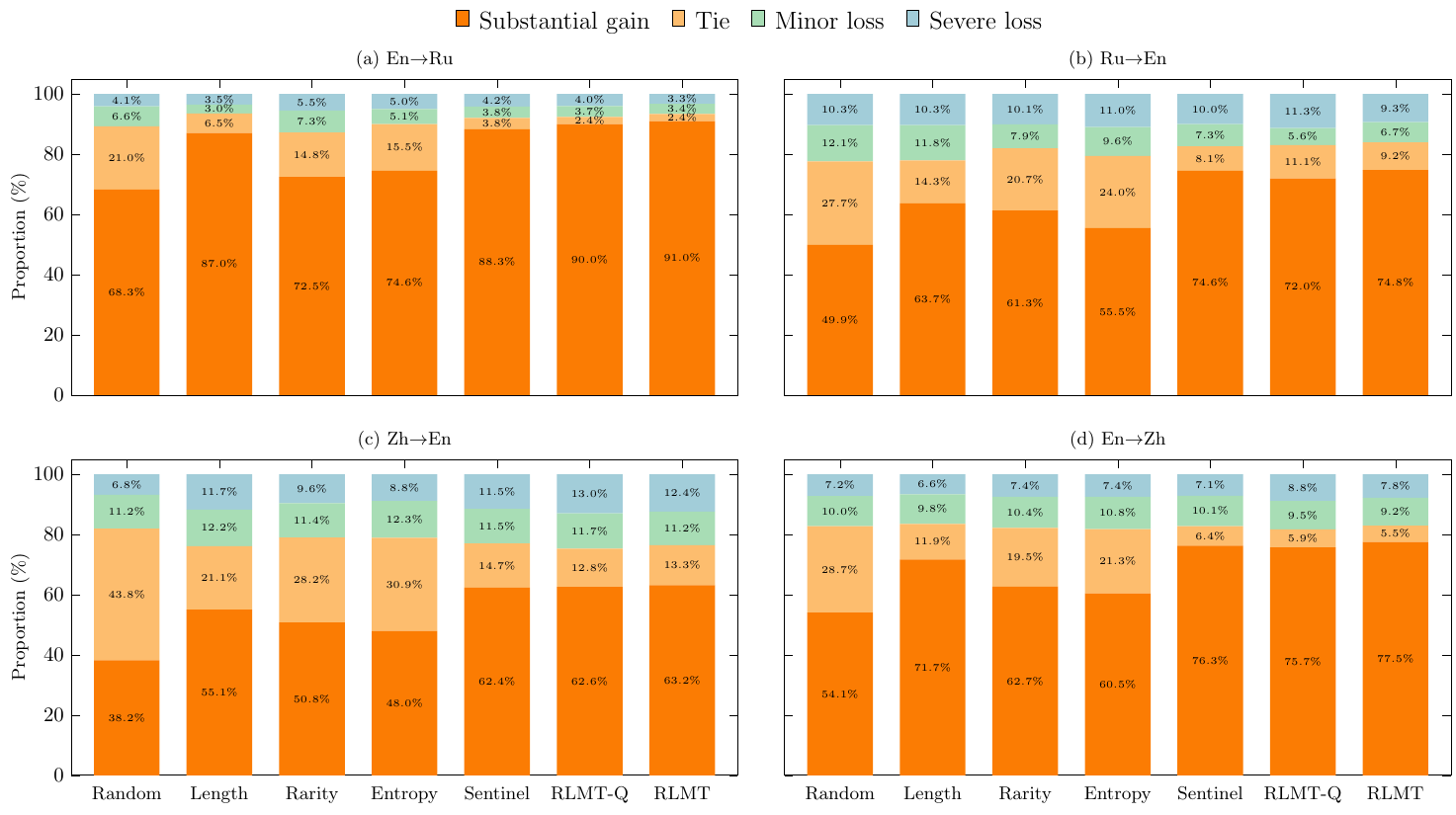}
\caption{Gain-bucket distribution among routed-to-large model requests under budget $p{=}0.3$.}
\label{fig:risk_all}
\end{figure*}

}

\end{document}